\documentclass{article}
\usepackage{spconf,amsmath,graphicx,hyperref,booktabs}
\DeclareUnicodeCharacter{04D9}{\rotatebox[origin=c]{180}{e}}
\graphicspath{{figures/}}

\title{GROUP RELATIVE POLICY OPTIMISATION IMPROVES MULTILINGUAL SPEECH RECOGNITION IN LOW-RESOURCE LANGUAGES}
\name{Tim Wenjie Hu, Evelyn Nafula Ouma, Benjamin Akera, John A. Quinn}
\address{Sunbird AI, Uganda}
\begin{document}
%
\maketitle
\begin{abstract}
Autoregressive encoder-decoder speech recognition models trained with teacher forcing can suffer from exposure bias, limiting their ability to recover from their own prediction errors at inference time. Recent work has shown that reinforcement learning with Group Relative Policy Optimisation (GRPO) improves speech recognition in major languages with LLM-based models. We extend this line of work to smaller, non-LLM encoder-decoder ASR models, using Whisper-large-v3 as a case study across 51 African languages. The benefits of this approach are most pronounced when training data is scarce or when input audio is noisy or degraded. Our resulting model, SunflowerASR, achieves the best published macro-averaged WER on the AfriVox-v2 (0.227) and SimbaBench (0.372) benchmarks, and is publicly available.
\end{abstract}
\section{Introduction}
\label{sec:intro}

Autoregressive speech recognition models such as Whisper~\cite{whisper} are typically trained only with supervised fine-tuning (SFT). This creates a risk of exposure bias: the model receives a training signal only for the case where it has been exactly correct about every preceding token, leaving it poorly prepared to recover from mistakes during inference.

Reinforcement learning methods provide an alternative, more representative of conditions at deployment time, where we generate entire rollouts from input audio sequences and then score each one as a whole.  Recent work has shown that Group Relative Policy Optimisation (GRPO)~\cite{grpo}, applied to a speech encoder attached to an Llama3 LLM decoder, can improve robustness on major European languages \cite{grpo_asr}, and is effective at reducing insertion errors such as hallucinations and repetition glitches with another LLM-based model, Qwen2-Audio-7B, in English~\cite{grounded_in_sound}. It has also been demonstrated that GRPO outperforms SFT in English when adapting a Llama-based ASR decoder using synthetic speech~\cite{kumar2026synthetic}.

In this work, we investigate whether these benefits extend to smaller, non-LLM encoder-decoder architectures, and to languages with scarce training data. We take the case of Whisper-large-v3~\cite{whisper} (1.55B parameters) and set up GRPO training with the objective of minimising character error rate (CER). We train a single model to transcribe 51 African languages, with per-language training data ranging from 5.2 hours (Ruruuli) to 1,392 hours (Kinyarwanda), and observe consistent improvements, which are most pronounced for noisy test audio and for languages with scarce training data. Our resulting model, SunflowerASR, achieves the best published average WER (per-language rankings vary) on the AfriVox-v2 and SimbaBench benchmarks, and can be accessed at \url{https://hf.co/Sunbird/SunflowerASR-51-african-languages}.

\section{Datasets and base model}
\label{sec:model_and_datasets}

\paragraph*{Training corpus}
\label{sec:corpus}
The training corpus combines 110 dataset subsets covering 51 African languages, a mean of 2.2 independent sources per language. Sources include Common Voice~\cite{commonvoice}, African Voices~\cite{africanvoices}, Waxal~\cite{waxal} and other publicly available datasets. Full source and language level details are available in our \href{https://huggingface.co/Sunbird/SunflowerASR-51-african-languages}{Hugging Face release}. After filtering it holds 3.82\,M clips totalling 7\,412 hours of speech. 
Training rows are filtered to remove suspected noisy labels, using an intermediate ASR model checkpoint, and then sample packed.

\begin{figure}[hb]
  \centering
  \includegraphics[width=\linewidth]{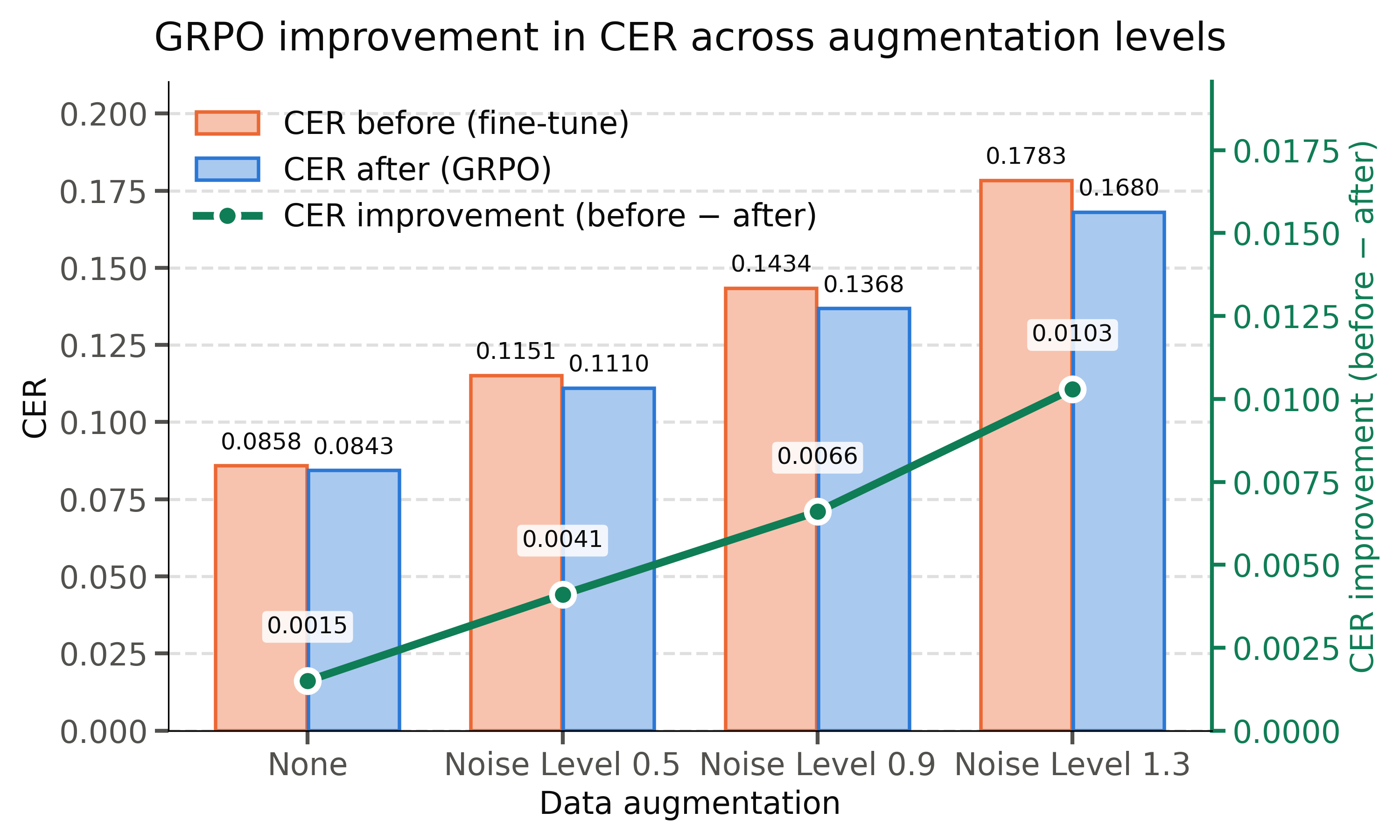}
  \caption{CER before (supervised fine-tuning only) and after GRPO for Whisper-large-v3 on 51 African languages. The improvement becomes more pronounced at increasing levels of background noise.}
  \vspace{-0.2cm}
  \label{fig:grpo_vs_augmentation}
\end{figure}

\paragraph*{Supervised fine-tuning}
\label{sec:sft}
We carry out full parameter fine tuning of the 1.55\,B parameters of the Whisper-large-v3 model from OpenAI~\cite{whisper}. Training runs for 5 epochs over the packed corpus with a per-device batch of 64 packed samples and 4 gradient accumulation steps, an effective batch of 256 packed samples, or roughly 1.7 hours of audio per optimiser step, giving approximately 4\,300 optimiser steps per epoch. Audio augmentation is applied on the fly, including additive noise drawn from a corpus of Ugandan ambient sound recordings, and modifications to speed and bandwidth. We use a peak learning rate of $2\times10^{-5}$ with 100 warmup steps and cosine decay, and label smoothing of $0.1$. 

\section{Reinforcement learning with GRPO}
\label{sec:grpo}

Supervised fine-tuning optimises the per-token likelihood of one reference transcript, so the model only ever receives a training signal for the case in which it has been exactly right about every preceding token. Group relative policy optimisation (GRPO) instead optimises the sequence-level metric we report. For a clip $x$ with reference transcript $y$, the policy $\pi_\theta$ samples a group of $G$ complete transcripts $\{\hat{y}_1,\dots,\hat{y}_G\}$, each is scored by a reward, and the group's own mean is used as the baseline against which its members are ranked:
\begin{equation}
r_i = -\min\big(\mathrm{CER}(\hat{y}_i, y),\, 1\big), \qquad
A_i = \frac{r_i - \mathrm{mean}(\mathbf{r})}{\mathrm{std}(\mathbf{r})} .
\label{eq:grpo_reward}
\end{equation}
Probability mass is then moved towards the rollouts with positive advantage, under a token-level normalised objective with a KL penalty of $\beta=0.01$ against the initial policy.


\paragraph*{Selecting examples with headroom}
\label{sec:selection}
In Equation~\ref{eq:grpo_reward}, advantages are defined relative to the group, so a clip whose $G$ rollouts all receive the same reward produces $A_i = 0$ for every member and contributes no gradient at all, however hard that clip is in absolute terms. Both failure modes are common in a 51-language corpus: clips the model already transcribes perfectly, and clips it fails on so completely that every rollout is equally wrong. GRPO can learn to make improvements for the clips on which the model is capable of a good transcript but does not reliably produce one.

To measure this directly, we use the SFT checkpoint in sampling mode over a random sample of the training data, generating $10$ transcripts per clip.
For each clip we then compute the headroom,
\begin{equation}
g = \mathrm{median}_i\ \mathrm{CER}(\hat{y}_i, y) \;-\; \min_i\ \mathrm{CER}(\hat{y}_i, y)\ ,
\label{eq:gap}
\end{equation}
the gap between the CER of a typical sample and that of the best sample in the group. A large $g$ means a better transcript is already inside the model's sampling distribution, and only needs to be made more probable. We keep the $200$ clips with the largest $g$ from each training data subset, totalling $15,285$ clips. Section~\ref{sec:ablations} ablates this choice against drawing the same number of clips per subset at random.


\paragraph*{Training}
\label{sec:grpo_training}
GRPO training is applied to the decoder of the model only, 906 M of the model's 1.55B parameters. We generate $G=10$ rollouts per training example, with 64 distinct clips per update for a batch size of 640 at peak learning rate $10^{-5}$. The batch is deliberately large for a run of this length. Because the training set is shuffled across all 51 languages, a batch of 64 clips draws on many languages at once, and a single update is therefore an average over them rather than a step that improves whichever language a small batch happened to contain. This is the main practical difference we found between the multilingual and the single-language case: with small batches, updates tend to improve one language at the expense of others. Section~\ref{sec:ablations} ablates the batch size directly. The Whisper encoder is frozen and only the decoder is trained, on the hypothesis is that GRPO is most effective at correcting decoding errors such as spelling, word choice and orthographic convention.



\section{Results}
\label{sec:results}



\begin{figure*}[t]
  \centering
  \includegraphics[width=\textwidth]{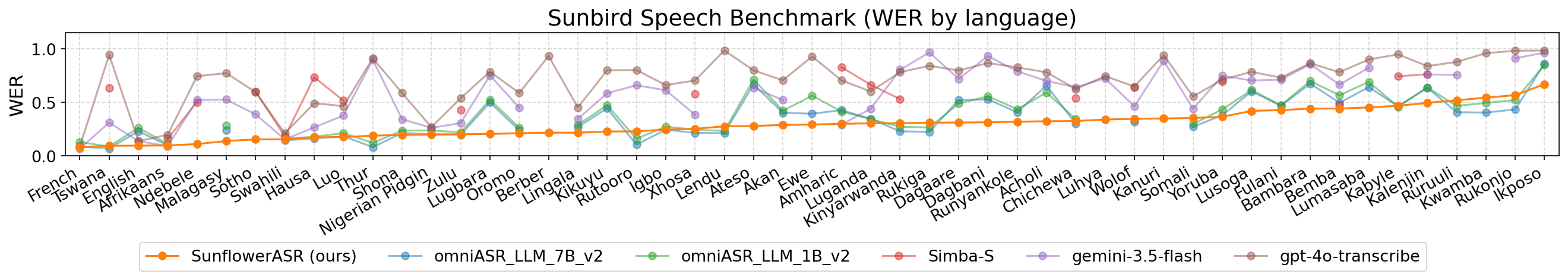}
  \caption{WER by language on the Sunbird Speech Benchmark, comparing SunflowerASR against external baselines. Languages are sorted by the WER of SunflowerASR.}
  \label{fig:sunbird_benchmark}
\end{figure*}

We evaluate SunflowerASR, our final GRPO model, on three benchmarks, shown per language in Fig.~\ref{fig:sunbird_benchmark} and Tables~\ref{tab:afrivox} and~\ref{tab:simbabench}. The first is our own Sunbird Speech Benchmark~\cite{sunbird_speech_benchmark_2026}, which we assemble from the held-out test splits of the 110 dataset subsets of the training corpus (Section~\ref{sec:corpus}), capped at 200 clips each. It scores $20,213$ clips over 51 languages drawn from 28 source corpora, and is the only one of the three that covers the full language range of the model. 
The second is AfriVox-v2~\cite{afrivox-v2}, an in-the-wild benchmark of $251$\,h aggregating the Africa Next Voices, Waxal and AfriVox v1 corpora over a 10-domain taxonomy spanning health, finance, government and agriculture; it is largely spontaneous rather than read, and we score $111,051$ clips in the 14 languages that have metrics reported in their paper. The third is SimbaBench~\cite{simbabench}, which aggregates read and conversational speech from Common Voice, BembaSpeech, Lwazi, NCHLT and regional collections. On all three benchmarks, our final model has superior accuracy to all published alternatives: see Figure \ref{fig:sunbird_benchmark} and Tables \ref{tab:afrivox} and \ref{tab:simbabench}.


\begin{table}[t]
  \centering
  \caption{WER by language on AfriVox-v2 for the 14 languages with published
  baselines, which report WER only. Simba-S is our own evaluation; it does not
  cover three of the languages, and its macro is over the other eleven.}
  \label{tab:afrivox}
  \vspace{2pt}
  \footnotesize
  \setlength{\tabcolsep}{3pt}
  \begin{tabular}{@{}l rrrrrrr@{}}
    \toprule
    & Sunflower & Sahara & Gemini & \multicolumn{3}{c}{Omni-CTC} & Simba \\
    \cmidrule(lr){5-7}
    Language & ASR (ours) & v2 & 3 Flash & 7B & 1B & 300M & S \\
    \midrule
    Swahili & 0.108 & \textbf{0.071} & 0.076 & 0.077 & 0.097 & 0.152 & 0.150 \\
    Tswana & \textbf{0.116} & 0.140 & 0.275 & 0.142 & 0.210 & 0.277 & 0.604 \\
    Kinyarwanda & 0.133 & \textbf{0.066} & 0.165 & 0.104 & 0.138 & 0.215 & 0.591 \\
    Sotho & \textbf{0.152} & 0.195 & 0.346 & 0.421 & 0.473 & 0.533 & 0.596 \\
    Luganda & \textbf{0.160} & 0.393 & 0.315 & 0.422 & 0.428 & 0.483 & 0.341 \\
    Zulu & \textbf{0.161} & 0.256 & 0.237 & 0.202 & 0.256 & 0.346 & 0.396 \\
    Xhosa & \textbf{0.166} & 0.206 & 0.292 & 0.200 & 0.256 & 0.344 & 0.501 \\
    Shona & \textbf{0.209} & 0.249 & 0.368 & 0.286 & 0.322 & 0.425 & --- \\
    Amharic & 0.251 & 0.253 & \textbf{0.249} & 0.327 & 0.374 & 0.490 & 0.716 \\
    Hausa & 0.288 & 0.285 & \textbf{0.269} & 0.502 & 0.366 & 0.401 & 0.835 \\
    Yoruba & 0.316 & \textbf{0.271} & 0.296 & 0.361 & 0.365 & 0.438 & 0.723 \\
    Fulani & \textbf{0.343} & 0.352 & 0.729 & 0.557 & 0.569 & 0.588 & --- \\
    Akan & 0.347 & \textbf{0.307} & 0.456 & 0.447 & 0.497 & 0.549 & --- \\
    Igbo & 0.425 & \textbf{0.287} & 0.425 & 0.459 & 0.395 & 0.446 & 0.998 \\
    \midrule
    Overall & \textbf{0.227} & 0.238 & 0.321 & 0.322 & 0.339 & 0.406 & 0.586 \\
    \bottomrule
  \end{tabular}
\end{table}

\begin{table}[tb]
  \centering
  \caption{WER by language on SimbaBench over the 19 languages in common. Baselines from the public leaderboard.}
  \label{tab:simbabench}
  \vspace{2pt}
  \footnotesize
  \setlength{\tabcolsep}{4pt}
  \begin{tabular}{@{}l rrrrr@{}}
    \toprule
    & SunflowerASR & OmniASR & Simba & Simba & Gemini \\
    Language & (Ours) & LLM-7\,B & S & M & 2.5 Pro \\
    \midrule
    Hausa & \textbf{0.167} & 0.169 & 0.641 & 0.291 & 0.219 \\
    Luganda & 0.220 & \textbf{0.171} & 0.227 & 0.357 & 0.421 \\
    Swahili & 0.246 & \textbf{0.156} & 0.163 & 0.252 & 0.168 \\
    Kinyarwanda & 0.251 & \textbf{0.192} & 0.531 & 0.367 & 0.463 \\
    Luo & 0.297 & \textbf{0.271} & 0.391 & 0.442 & 0.494 \\
    Igbo & \textbf{0.304} & 0.362 & 0.777 & 0.850 & 0.428 \\
    Afrikaans & 0.312 & 0.279 & \textbf{0.145} & 0.313 & 0.158 \\
    Wolof & \textbf{0.324} & 0.583 & 0.581 & 0.586 & 1.830 \\
    Yoruba & 0.335 & 0.232 & \textbf{0.201} & 0.415 & 0.524 \\
    Kabyle & 0.349 & \textbf{0.287} & 0.604 & 0.643 & 0.933 \\
    Tswana & 0.353 & 0.489 & \textbf{0.187} & 0.306 & 0.464 \\
    Bemba & \textbf{0.374} & 0.392 & 0.391 & 0.437 & 0.635 \\
    Sotho & 0.441 & 0.634 & \textbf{0.210} & 0.327 & 0.495 \\
    Ndebele & 0.444 & 0.588 & \textbf{0.246} & 0.358 & 0.413 \\
    Zulu & 0.459 & 0.401 & 0.585 & \textbf{0.333} & 0.359 \\
    Xhosa & 0.466 & 0.448 & \textbf{0.272} & 0.383 & 0.466 \\
    Chichewa & 0.545 & 0.609 & \textbf{0.223} & 0.468 & 0.428 \\
    Kalenjin & \textbf{0.580} & 0.700 & 0.713 & 0.734 & 0.959 \\
    Amharic & 0.600 & \textbf{0.234} & 0.330 & 0.525 & 0.457 \\
    \midrule
    Overall & \textbf{0.372} & 0.379 & 0.390 & 0.442 & 0.543 \\
    \bottomrule
  \end{tabular}
\end{table}

To evaluate the benefits of GRPO specifically, compared to the SFT-only initial checkpoint, we create three additional versions of 51-language evaluation set, with noise drawn from a corpus of Ugandan street recordings at three mean relative amplitudes, $0.5$, $0.9$ and $1.3$, and also score the undegraded audio as a control. 
CER is consistently better after GRPO, and the improvement grows monotonically with the severity of the degradation, as shown in Fig.~\ref{fig:grpo_vs_augmentation}. 



\begin{figure}[t]
  \centering
  \hspace{-0.2cm}\includegraphics[width=\linewidth]{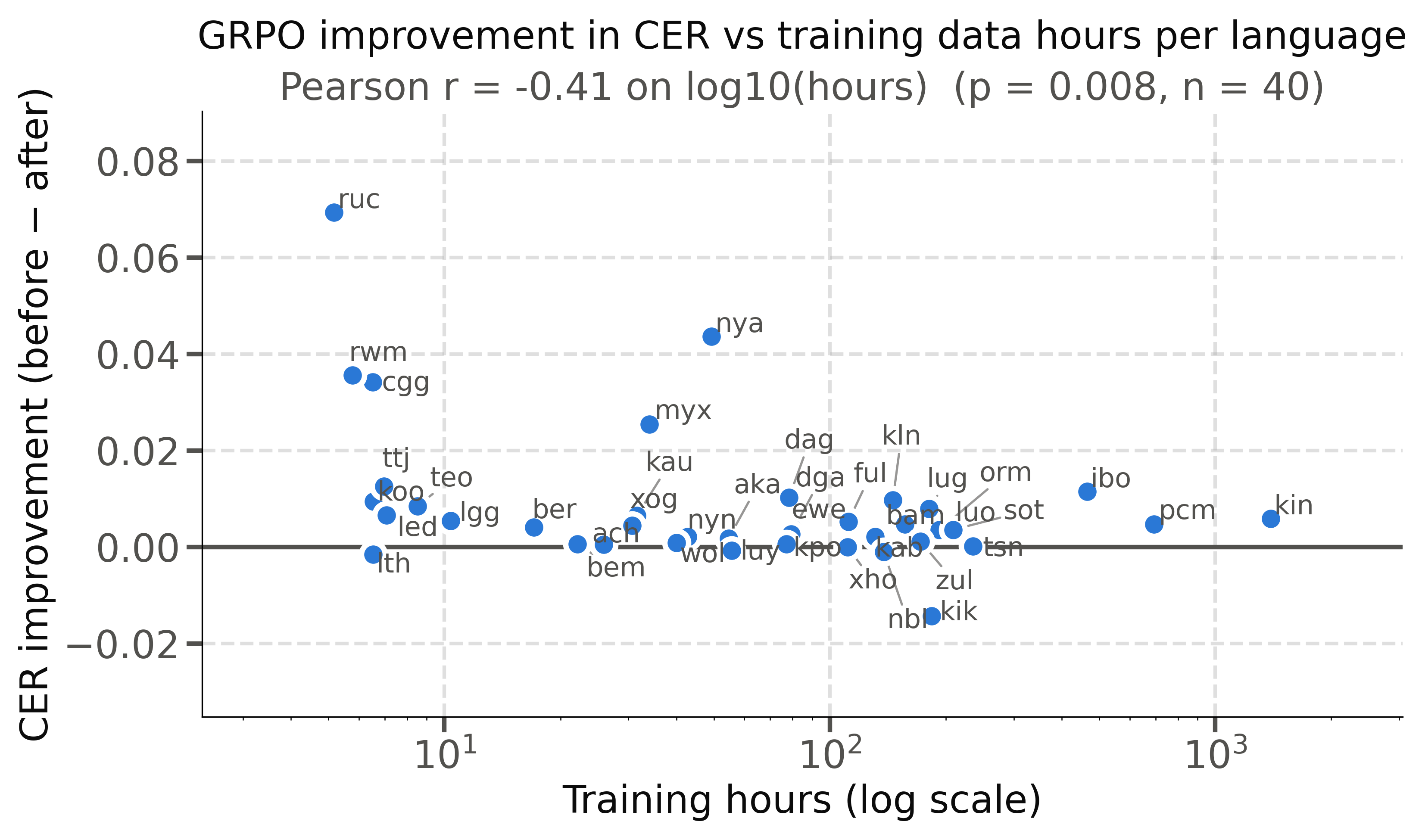}
  \caption{Per-language CER improvement from GRPO against the number of hours of supervised training data for that language.}
  \label{fig:hours_vs_cer_diff}
\end{figure}

GRPO mainly repairs failures under acoustic stress, and the low-resource languages with the less supervised data gain the most. Fig.~\ref{fig:hours_vs_cer_diff} illustrates each language's CER improvement against the hours of supervised audio it contributed to the training corpus, for which there is a correlation of $r=-0.41$ on $\log_{10}$ hours, with 35 of 40 languages improving.



\paragraph*{Practical deployment audio}
\label{sec:practical_eval}

Table~\ref{tab:practical_eval} reports the effect of GRPO on Sunbird's \emph{salt-practical-eval} set, a 60-clip validation set assembled to reflect real deployment. The recordings were collected in Uganda and cover Ugandan English (28 clips) and Luganda (32 clips) across three deployment settings: citizen feedback gathered outside hospitals and government offices, conversations recorded at the call centre of a bank, and phone-in contributions to an interactive radio programme. The audio is spontaneous and conversational, captured in noisy public spaces, with overlapping turns, hesitation and backchannel markers, and frequent code-switching from Luganda into English for domain vocabulary and numerals. GRPO improves both metrics in both languages more significantly than for benchmark test data: the aggregate reduction is $6.2\%$ relative on WER and $9.7\%$ relative on CER. 

\subsection{Ablations}
\label{sec:ablations}

Three ablations isolate the contribution of the reinforcement objective itself, of the clip selection of Section~\ref{sec:selection}, and of the batch size of Section~\ref{sec:grpo_training}; results are in Table~\ref{tab:ablations}.

\textbf{GRPO against further supervised training on the same clips.} GRPO adds an extra epoch of training over a curated $15,285$-clip subset, so its gain could in principle come from that extra pass rather than from the reinforcement signal. We test this by continuing \emph{supervised} fine-tuning from the same checkpoint for one further epoch over exactly the clips GRPO was given. This regresses the model on both WER and CER (Table~\ref{tab:ablations}). The same data is thus useful under a reinforcement objective and harmful under a supervised one. 


\textbf{Selecting examples with headroom.} Replacing the headroom ranking of Equation~\ref{eq:gap} with $200$ clips drawn uniformly at random from each subset, everything else held fixed, gives WER $0.4220$ and CER $0.1714$. Random selection still improves on the supervised checkpoint, but it recovers only about two thirds of the CER gain, $-3.9\%$ against $-5.8\%$ relative. 

\textbf{Batch size.} Reducing the batch from $64$ distinct clips to $16$, at the same $G=10$ and over the same selected clips, gives WER $0.4243$ and CER $0.1689$. The regression is smaller than for the other two ablations and is clearer on WER ($-0.8\%$ against $-1.8\%$ relative) than on CER, but its direction matches the argument of Section~\ref{sec:grpo_training}: with $51$ languages shuffled together, $16$ clips is a thin sample of the language set, and an update is correspondingly likely to benefit only the languages that batch happened to contain.

\begin{table}[]
  \centering
  \caption{Effect of GRPO on the Sunbird \emph{salt-practical-eval} set, by deployment scenario and language.} 
  \label{tab:practical_eval}
  \vspace{2pt}
  \footnotesize
  \setlength{\tabcolsep}{2pt}
  \begin{tabular}{@{}lr rrr rrr@{}}
    \toprule
    & & \multicolumn{3}{c}{WER} & \multicolumn{3}{c}{CER} \\
    \cmidrule(lr){3-5} \cmidrule(lr){6-8}
    Scenario & $n$ & Before & After & $\Delta$ & Before & After & $\Delta$ \\
    \midrule
    \multicolumn{8}{@{}l}{\emph{Ugandan English}} \\
    \quad Citizen feedback & 12 & 0.324 & 0.282 & $-13.0\%$ & 0.223 & 0.182 & $-18.0\%$ \\
    \quad Call centre      & 16 & 0.302 & 0.273 & $-9.4\%$  & 0.213 & 0.182 & $-14.8\%$ \\
    \addlinespace
    \multicolumn{8}{@{}l}{\emph{Luganda}} \\
    \quad Citizen feedback & 8  & 0.705 & 0.683 & $-3.0\%$ & 0.324 & 0.317 & $-2.2\%$ \\
    \quad Radio phone-in   & 15 & 0.535 & 0.528 & $-1.4\%$ & 0.140 & 0.132 & $-5.4\%$ \\
    \quad Call centre      & 9  & 0.602 & 0.565 & $-6.2\%$ & 0.272 & 0.265 & $-2.5\%$ \\
    \midrule
    All & 60 & 0.423 & 0.397 & $-6.2\%$  & 0.215 & 0.194 & $-9.7\%$  \\
    \bottomrule
  \end{tabular}
\end{table}

\begin{table}[]
  \centering
  \caption{Ablations on the noise-$1.3$ set ($20,347$ rows), with all variants starting from the same SFT checkpoint.}
  \label{tab:ablations}
  \vspace{2pt}
  \footnotesize
  \setlength{\tabcolsep}{3pt}
  \begin{tabular}{@{}l rrrr@{}}
    \toprule
    Variant & WER & $\Delta$ & CER & $\Delta$ \\
    \midrule
    SFT checkpoint (initialisation) & 0.4279 & --- & 0.1783 & --- \\
    $+$\,1 further SFT epoch & 0.4794 & $+12.0\%$ & 0.2089 & $+17.2\%$ \\
    GRPO, random clips per subset & 0.4220 & $-1.4\%$ & 0.1714 & $-3.9\%$ \\
    GRPO, 16 clips per batch & 0.4243 & $-0.8\%$ & 0.1689 & $-5.3\%$ \\
    GRPO, full (SunflowerASR) & \textbf{0.4201} & $-1.8\%$ & \textbf{0.1680} & $-5.8\%$ \\
    \bottomrule
  \end{tabular}
\end{table}


\begin{table}
  \centering
  \caption{Substitution quality on the $20,124$ rows where neither model loops.}
  \label{tab:subquality}
  \vspace{2pt}
  \footnotesize
  \setlength{\tabcolsep}{4pt}
  \begin{tabular}{@{}lrr@{}}
    \toprule
    Measure & SFT & GRPO \\
    \midrule
    Word substitutions & 114,755 & 115,272 \\
    Character substitutions & 165,096 & \textbf{160,220} \\
    Mean char.\ edit distance per substitution & 3.437 & \textbf{3.385} \\
    Substitutions within $\le1$ edit & 26.7\% & \textbf{27.4\%} \\
    Substitutions within $\le2$ edits & 44.9\% & \textbf{45.8\%} \\
    Total char.\ cost of substitutions & 394,368 & \textbf{390,231} \\
    Long words cut to $\le$ half their length & 18.8\% & \textbf{18.0\%} \\
    \bottomrule
  \end{tabular}
\end{table}

\begin{table}
  \centering
  \caption{Cases where the supervised model heard a fragment or an unrelated word and GRPO heard a near-miss spelling.}
  \label{tab:nearmiss}
  \vspace{2pt}
  \scriptsize
  \setlength{\tabcolsep}{3pt}
  \begin{tabular}{@{}llllr@{}}
    \toprule
    Lang. & Reference & SFT heard & GRPO heard & Edits \\
    \midrule
    Oromo & libiraal & liibiraaldimookiraasiitti & liibiraal & 17\,$\to$\,1 \\
    Rukonjo & ngerikulhakulhania & ebyo & erikulhakulhania & 17\,$\to$\,2 \\
    Shona & vobva & rwakachenerukira & wobva & 15\,$\to$\,1 \\
    Luganda & bibili & ebitakyalabirwaako & bibiri & 14\,$\to$\,1 \\
    Ruruuli & ebyankulakulana & mbe & ebyangulagulana & 14\,$\to$\,2 \\
    Kanuri & kәnasartiyende & dә & kәnasartiyande & 13\,$\to$\,1 \\
    Zulu & amabhayisikili & i & mubhayisikili & 13\,$\to$\,2 \\
    Rutooro & baali & batarukwegesaana & baaba & 13\,$\to$\,2 \\
    Somali & dheellitirka & wax & dheelitirka & 12\,$\to$\,1 \\
    Kwamba & ndia & ndibhakazwisikilangu & ndiabho & 16\,$\to$\,3 \\
    \bottomrule
  \end{tabular}
\end{table}

\subsection{Characterisation of improvements}
\label{sec:discussion}

To understand the nature of the improvements, we aligned the $20,347$ hypotheses of the noise-$1.3$ set against the references, and observed two different mechanisms. 
The first is the suppression of degenerate decoding. Under acoustic stress the supervised model sometimes falls into a repetition loop and emits hundreds of words for a short utterance. 
There are 223 rows ($1.1\%$ of the evaluation) for which the SFT-only model becomes stuck in a loop. GRPO repairs these decoding failures, and this accounts for the WER gain.

On the remaining $20,124$ rows the two models are wrong about equally often; word substitutions are flat at $+0.5\%$, yet character substitutions fall by $3.0\%$. 
The characteristic supervised failure is to replace a word either with a fragment or with an unrelated long compound, while GRPO returns a near-miss spelling that tracks the acoustics through to the end of the word (Table~\ref{tab:nearmiss}). 


\section{Conclusion}
\label{sec:conclusion}

We have shown that GRPO transfers from LLM-decoder systems to a smaller non-LLM encoder-decoder model, in a multilingual, low-resource and noisy setting. The benefits increase for languages with fewer data resources (e.g. tens of hours of audio), and in test time conditions with challenging audio. Selection of training examples with headroom and appropriate batch sizing are helpful to attain improvements in accuracy. This therefore appears to be a widely generalisable post-training method for autoregressive ASR models.

\bibliographystyle{IEEEbib}
\bibliography{strings,refs}

\end{document}